\documentclass[10pt,twocolumn,letterpaper]{article}

\usepackage[pagenumbers]{iccv} 
\usepackage{dsfont}
\usepackage[hang,flushmargin]{footmisc}
\definecolor{iccvblue}{rgb}{0.21,0.49,0.74}
\usepackage[pagebackref,breaklinks,colorlinks,allcolors=iccvblue]{hyperref}

\def\paperID{5621} 
\def\confName{ICCV}
\def\confYear{2025}

\newcommand{\sysname}{WP-LoRA}
\title{From Wardrobe to Canvas: Wardrobe Polyptych LoRA \\ for Part-level Controllable Human Image Generation}

\newcommand*{\affmark}[1][*]{\textsuperscript{#1}}
\newcommand\blfootnote[1]{%
  \begingroup
  \renewcommand\thefootnote{}\footnote{#1}%
  \addtocounter{footnote}{-1}%
  \endgroup
}

\author{
Jeongho Kim\affmark[1,2]$^\text{*}$ \; Sunghyun Park\affmark[1] \; Hyoungwoo Park\affmark[1] \; Sungrack Yun\affmark[1] \; \\ 
Jaegul Choo\affmark[2] \; Seokeon Choi\affmark[1]\; \vspace{0.12cm}\\
\affmark[1]Qualcomm AI Research$^\dagger$ \; \affmark[2]Korea Advanced Institute of Science and Technology (KAIST)\\
\texttt{\footnotesize\{jeonghok, sunpar, hwoopark, sungrack, seokchoi\}@qti.qualcomm.com} \\
\texttt{\footnotesize\{rlawjdghek, jchoo\}@kaist.ac.kr} \;
}

\begin{document}

\blfootnote{$^*$Work done during an internship at Qualcomm AI Research.}
\blfootnote{$^\dagger$ Qualcomm AI Research is an initiative of Qualcomm Technologies, Inc.}


\begin{figure}
\twocolumn[{
\renewcommand\twocolumn[1][]{#1}
\maketitle
\begin{center}
    \centering 
    \vspace{-5mm}
    \includegraphics[width=0.92\linewidth]{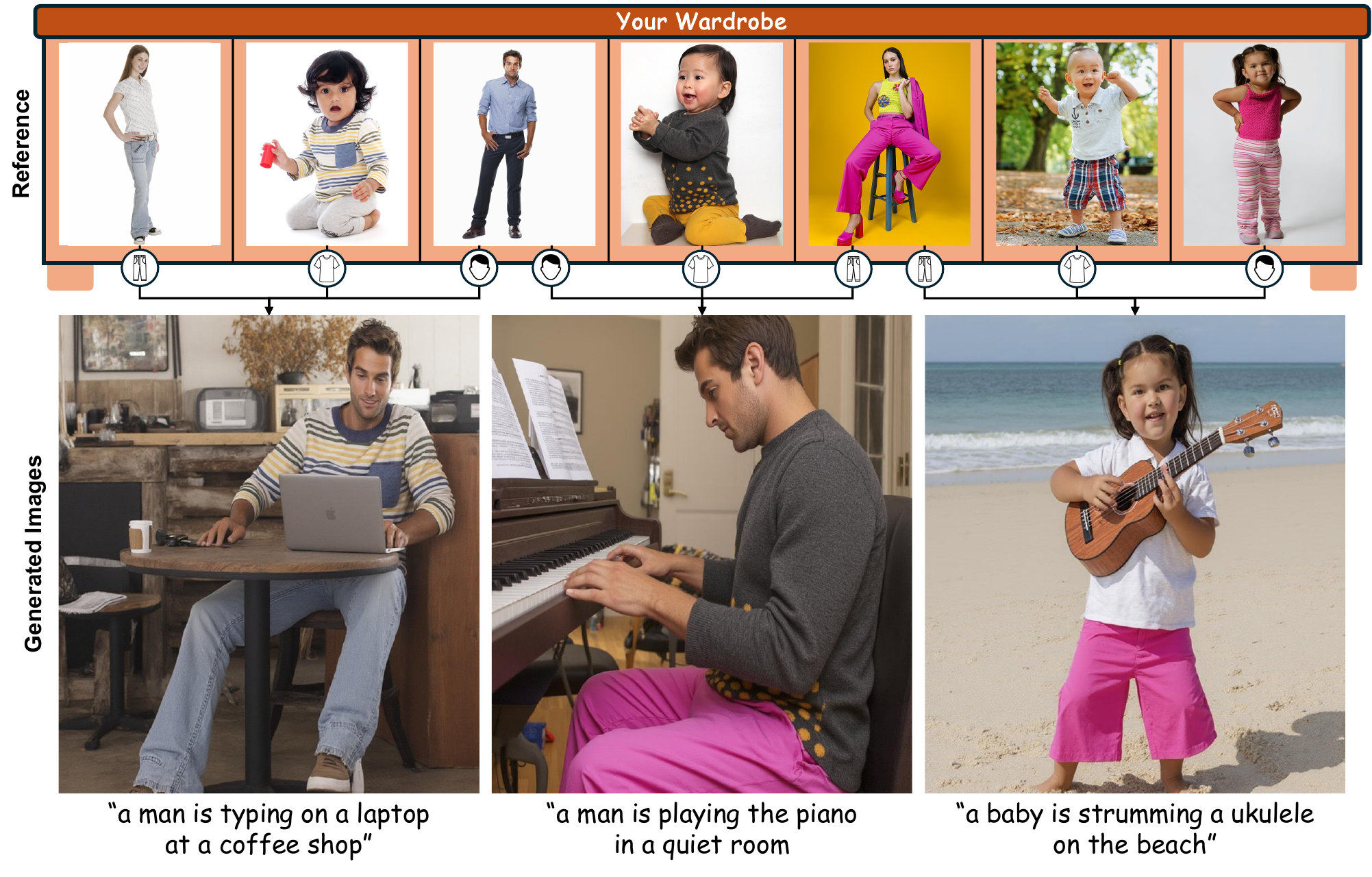}
    \vspace{-2mm}
    \captionof{figure}{Wardrobe Polyptych LoRA is a framework for part-level controllable human image generation, enabling the composition of individuals (\eg face, upper clothing, and lower clothing) present in the training data while also generalizing to unseen individuals without requiring additional fine-tuning. The source individuals depicted in this figure are from our Persona-36 dataset.}
    \label{fig:teaser}
\end{center}
}]
\end{figure}


    \vspace{-7mm}
\begin{abstract}
    Recent diffusion models achieve personalization by learning specific subjects, allowing learned attributes to be integrated into generated images. 
    However, personalized human image generation remains challenging due to the need for precise and consistent attribute preservation (e.g., identity, clothing details). 
    Existing subject-driven image generation methods often require either (1) inference-time fine-tuning with few images for each new subject or (2) large-scale dataset training for generalization. Both approaches are computationally expensive and impractical for real-time applications.
    To address these limitations, we present Wardrobe Polyptych LoRA, a novel part-level controllable model for personalized human image generation. 
    By training only LoRA layers, our method removes the computational burden at inference while ensuring high-fidelity synthesis of unseen subjects.
    Our key idea is to condition the generation on the subject's wardrobe and leverage spatial references to reduce information loss, thereby improving fidelity and consistency. 
    Additionally, we introduce a selective subject region loss, which encourages the model to disregard some of reference images during training.
    Our loss ensures that generated images better align with text prompts while maintaining subject integrity.
    Notably, our Wardrobe Polyptych LoRA requires no additional parameters at the inference stage and performs generation using a single model trained on a few training samples.
    We construct a new dataset and benchmark tailored for personalized human image generation. 
    Extensive experiments show that our approach significantly outperforms existing techniques in fidelity and consistency, enabling realistic and identity-preserving full-body synthesis.
\end{abstract}

\section{Introduction}
Human image generation—especially the task of composing different body parts (\eg, combining upper and lower body clothing) across various individuals—remains a significant challenge in computer vision applications such as virtual try-on and personalized human content generation~\cite{han2018viton,kim2024stableviton,li2024cosmicman}. 
Recent advances in diffusion models have shown remarkable success in image synthesis, with considerable progress by fine-tuning these models on specific subjects~\cite{ruiz2023dreambooth,avrahami2023break,kumari2023multi,gal2022image}. 
However, these per-subject optimization techniques (whether single- or multi-subject approaches) have a notable drawback: they require fine-tuning at inference time for each new subject. 
This makes them computationally expensive and impractical for real-time applications, particularly when new subjects need to be introduced.

In response to these challenges, there has been growing interest in methods that generate subject-driven images without requiring per-subject optimization~\cite{huang2024parts,xu2024magicanimate,li2023blip,ye2023ip}.
However, these approaches still exhibit several limitations. 
First, they often rely on an additional image encoder for subject-specific information, which increases the overall parameter count and leads to high computational overhead. 
Second, they typically depend on large-scale datasets for pre-training, which necessitates extensive data collection and results in long, resource-intensive training processes. 
Third, they struggle to reflect multiple conditional images simultaneously.
These factors make them less practical for real-world applications and underscore the difficulty of achieving computationally efficient, scalable, and data-efficient full-body generation in data-constrained environments.



Recent studies have shown that large-scale text-to-image diffusion models based on diffusion transformers~\cite{peebles2023scalable} have the capability to transfer information from conditional images through attention mechanisms.
In-context LoRA~\cite{huang2024context} trains LoRA layers that learn consistent relationships between concatenated images using only a few data samples (20-100 images), enabling the generation of images with consistent appearances or styles, such as portrait illustrations.
Diptych prompting~\cite{shin2024large} allows FLUX-based inpainting models~\cite{alimama2024flux} to perform personalized image generation by conditioning a single subject through attention.
However, research on incorporating multiple conditional images remains underexplored.

In this study, we propose~\sysname~(Wardrobe Polyptych LoRA), a novel framework for human image generation, which composites multiple conditional images including face, upper, and lower clothing.
WP-LoRA learns to transfer condition information through attention, generating images on a canvas that reflects text prompts from a collection of multiple conditional images, referred to as the wardrobe region.
By conditioning on this spatial region, our approach reduces spatial information loss and ensures greater consistency in the generated images without additional parameters.
Built on the FLUX inpainting model~\cite{black2024fluxfill}, WP-LoRA can be trained with only a few samples (100 images). 
Since simultaneously incorporating multiple conditional images remains a challenging task, we introduce a selective subject region loss to capture details from all input conditions.
This loss is applied only to specific subject regions during training, improving text prompt adherence across diverse poses and viewpoints, ensuring more accurate and realistic human image generation.


Our method requires no additional parameters at the inference stage and allows for the composition of multiple individuals to form new identities. Despite being trained on a limited dataset,~\sysname~generalizes effectively to unseen identities without further fine-tuning. To rigorously evaluate our approach, we construct a new dataset called Persona-36 and benchmark tailored to this task. Extensive experiments reveal that~\sysname~substantially outperforms existing personalization methods in both fidelity and diversity, offering identity-consistent part-level controllability.

In summary, our key contributions are as follows:
\begin{itemize}
    \item We propose a novel framework that conditions on structured wardrobe region, preserving spatial information from reference images, and enhancing consistency and accuracy in full-body synthesis.

    \item We introduce a selective subject region loss, which randomly drops the reference subjects during training, resulting in better adherence to the subject's appearance across various poses and viewpoints.

    \item Our approach accurately incorporates unseen identities without additional fine-tuning and supports multi-individual composition for synthesizing new identities—all without any extra parameters at inference.
\end{itemize}

\begin{figure*}
    \centering
    \includegraphics[width=\linewidth]{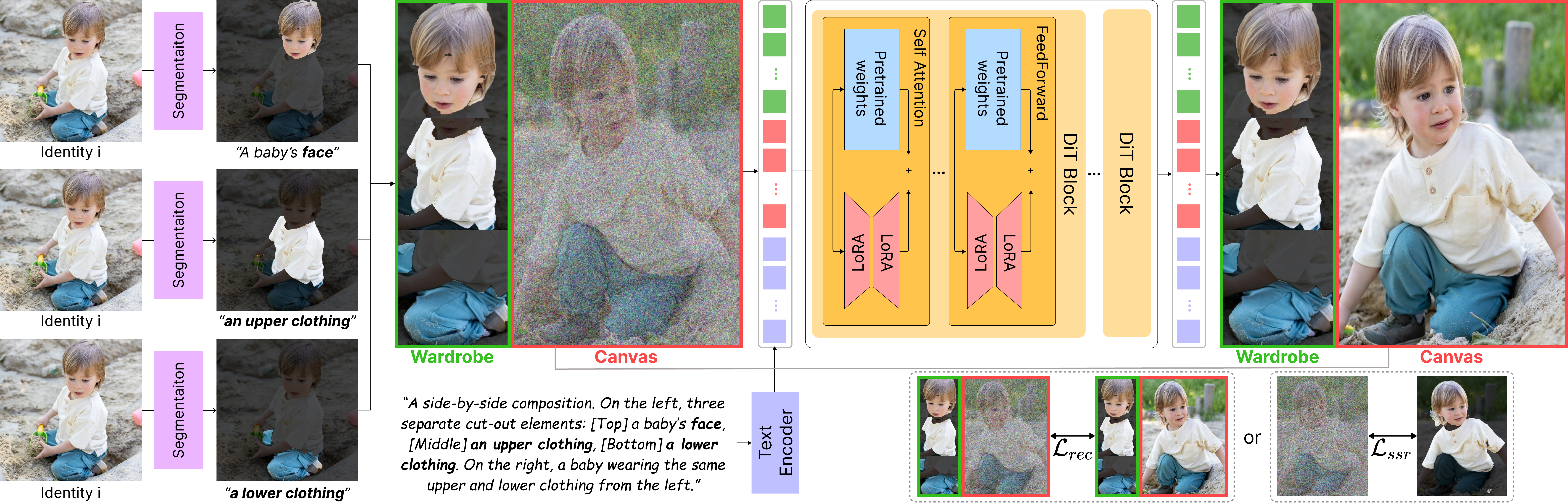}
    \caption{The overall structure of our methodology for part-level controllable human generation. When multiple subjects from a given identity are provided, we segment each subject and place their corresponding regions into a specific region of the generated image (wardrobe). Textual descriptions representing each subject's category are combined with wardrobe and canvas tokens as input to the DIT model. Only LoRA layers are trained to reconstruct the same identity within the canvas region and are shared across all identities. During training, we stochastically select either the reconstruction loss or the proposed selective subject region loss at each training step.}
    \vspace{-0.3cm}
    \label{fig:main}
\end{figure*}

\section{Related Work}
\noindent\textbf{Subject-driven Image Generation.}
Subject-driven image generation~\cite{ruiz2023dreambooth,ye2023ip,alaluf2023neural} has emerged as a prominent research field creating images based on user-specified subjects. 
In particular, recent personalization~\cite{ruiz2023dreambooth,kumari2023multi} techniques have achieved remarkable generation results requiring a few images for fine-tuning the generative models per subjects. 
They have been developed to either map specific subjects to unique tokens through inversion~\cite{gal2022image,avrahami2023break,kumari2023multi,alaluf2023neural,voynov2023p+} or learn specialized layers for each subject~\cite{ruiz2023dreambooth,yang2024lora}. 
These approaches have evolved from handling single subjects to enabling multi-subject composition, where tokens or layers learned for separate subjects can be combined to create complex scenes~\cite{gu2023mix,avrahami2023break,kong2024omg,yang2024lora}. A notable limitation of these methods is their requirement for new fine-tuning whenever a novel subject is introduced. 
This constraint stems from their lack of prior knowledge about unseen subjects, resulting in computational costs that make them impractical for real-time applications. 
In contrast, our approach does not focus on learning specific subjects; instead, we train the model to reference and compose subjects at designated locations, achieving high fidelity even for novel subjects that were not seen during training.

\noindent\textbf{Part-level Controllable Image Generation.}
With the remarkable image generation capabilities of large vision models~\cite{rombach2022high, peebles2023scalable, esser2024scaling}, recent studies have explored techniques for decomposing a single subject into distinct parts for creative composition and editing. 
Existing approaches~\cite{ng2024partcraft, huang2024parts} primarily focus on predefined category-based representations rather than subject-specific decompositions, leveraging these categories for composition.
For instance, PartCraft~\cite{ng2024partcraft} learns category-specific tokens to map parts of a reference image into the text modality, while Parts2Whole~\cite{huang2024parts} employs an additional image encoder to encode part-level reference subjects and conditions them via shared attention mechanism~\cite{hu2024animate,xu2024magicanimate,hertz2024style}. 
Although methods using external encoder preserve details by injecting features into the attention mechanism as keys and values, they can introduce a distribution shift between denoised samples—which retain residual noise—and the clean features from the external encoder. 
Furthermore, the additional encoders proposed in previous works~\cite{ye2023ip,li2023blip,huang2024parts} significantly increase memory consumption, thereby constraining the scalability of backbone models.
In contrast, our method leverages regions of the input image as conditioning areas for the reference subjects, eliminating the need for additional encoders. It demonstrates the ability to adapt and learn part-level control capabilities for novel subjects, even with fewer than 100 training images.

\section{Method}
Given images of multiple individuals, our objective is to compose a new person by referencing part-level subjects (e.g., face, upper clothing) from each individual. Instead of using a separate encoder to process these extracted subjects, we place them in a fixed “wardrobe” region of the generated image. The model then generates a novel image in the target “canvas” area by referencing these subjects in the wardrobe. Through this training strategy, the model learns to adapt to unseen individuals at test time: as long as the input part belongs to a known category, it accurately transfers its appearance into the final output.

\subsection{Preliminaries}
\noindent\textbf{Diffusion Transformers.}
Latent Diffusion Transformer (DiT)~\cite{peebles2023scalable,esser2024scaling} replaces or augments the conventional U-Net backbone in a diffusion model with a transformer architecture, leveraging self-attention layers to capture global context at each generation step. For memory efficiency, a variational autoencoder (VAE) compresses images into a latent representation. Within this latent space, DiT projects both image and text features into token embeddings, which are then refined by self-attention and feed-forward layers in each DiT block. By modeling long-range dependencies through attention, DiT often yields more coherent results than purely convolutional pixel-space diffusion approaches, efficiently capturing interactions across distant regions in the latent space.

\noindent\textbf{Flow Matching.}
Flow matching~\cite{lipman2022flow,esser2024scaling} aligns the flow from the noise to data distributions by optimizing a velocity field that gradually transforms noise into data over time. This ensures that the generative model learns a structured mapping from the noise distribution to the true data distribution. The flow matching loss is formulated as follows:
\begin{equation}
\mathcal{L}_{\mathrm{CFM}} = \mathbb{E}_{t,\,p_t(\mathbf{z}\mid \boldsymbol{\epsilon}),\,p(\boldsymbol{\epsilon})}
\Bigl[
\|\mathbf{v}_\Theta(\mathbf{z}, t) - \mathbf{u}_t(\mathbf{z}\mid \boldsymbol{\epsilon})\|^2
\Bigr],
\end{equation}
where $\mathbf{v}_\Theta(\mathbf{z}, t)$ denotes the velocity field parameterized by the neural network, and $\mathbf{u}_t(\mathbf{z}\mid \boldsymbol{\epsilon})$ is the conditional vector field describing the probabilistic path between noise and the true data distribution. The expectation $\mathbb{E}$ is taken over time $t$, the sample variable $\mathbf{z}$, and noise $\boldsymbol{\epsilon}$, thereby averaging the squared differences across all instances to provide a reliable measure of the model's generative capability.

\subsection{Wardrobe Polyptych LoRA}
We propose a novel framework for part-level controllable human generation. Suppose we have $n$ images for the same identity, $\{x_{1}, x_{2}, \dots, x_{n}\}$. Each image $x_{i}$ includes $m$ part-level subjects, represented as $s_{i} = \{s_{i1}, \dots, s_{im}\}$.
Whereas many existing methods convert multiple subjects into text-based embeddings (potentially losing spatial information), our approach preserves spatial detail by placing these subjects within a designated wardrobe region in the generated image.

As illustrated in Fig.~\ref{fig:main}, we position the $m$ subjects on the left side of the canvas region (here, $m=3$). Since each subject consisting of individuals occupies a small portion of the entire image, we design the wardrobe region at the same height as the canvas to align with the original rectangular layout. Additionally, to disentangle the target subject from the background or other nearby subjects, any non-target subject areas are masked out with the segmented masks, and we concatenate the wardrobe and canvas along the token dimension.

We design three types of text prompts: 1) a global prompt describing the overall structure of the image, 2) part-level prompts specifying each subject in the wardrobe region, and 3) a final composition prompt describing the generated image. Notably, these prompts do not contain unique texture or identity information, allowing them to remain unchanged for subjects within the same category (e.g., "face" for any face). Finally, the wardrobe and canvas images, along with their corresponding text prompts, are passed through a VAE and text encoder. The resulting latent representations are concatenated and serve as input to the DiT model:

\begin{equation}
    \begin{split}
        \mathbf{z}^{\prime} &= \mathcal{E}(\text{concat}(\{s_{t_j, j} \mid j=1,...,m; t_j \neq i\}, x_i)) \\
        \mathcal{L}_{\text{rec}} &= \mathbb{E}_{t,\,p_t(\mathbf{z}^{\prime} \mid \boldsymbol{\epsilon}),\,p(\boldsymbol{\epsilon})}
        \left[\|\mathbf{v}_\Theta(\mathbf{z}^{\prime}, t, \mathbf{m}) - \mathbf{u}_t(\mathbf{z}^{\prime} \mid \boldsymbol{\epsilon})\|^2        \right],
    \end{split}
\end{equation}
where $\mathcal{E}$ is the VAE encoder, and $\mathbf{m}$ is a mask representing the wardrobe and canvas regions.

We adapt and train LoRA~\cite{hu2022lora} layers within the DiT blocks; with the trained weights shared across all identities.
Instead of training subject-specific tokens~\cite{gal2022image,avrahami2023break} or LoRA weights~\cite{ruiz2023dreambooth}, our LoRA layers adapt to the proposed wardrobe-canvas template through reconstruction loss. 
This adaptation process enables the model to reference the fixed wardrobe region via attention, facilitating the generation of new individuals while maintaining detailed consistency.


\begin{figure*}[t!]
    \centering
    \includegraphics[width=\linewidth]{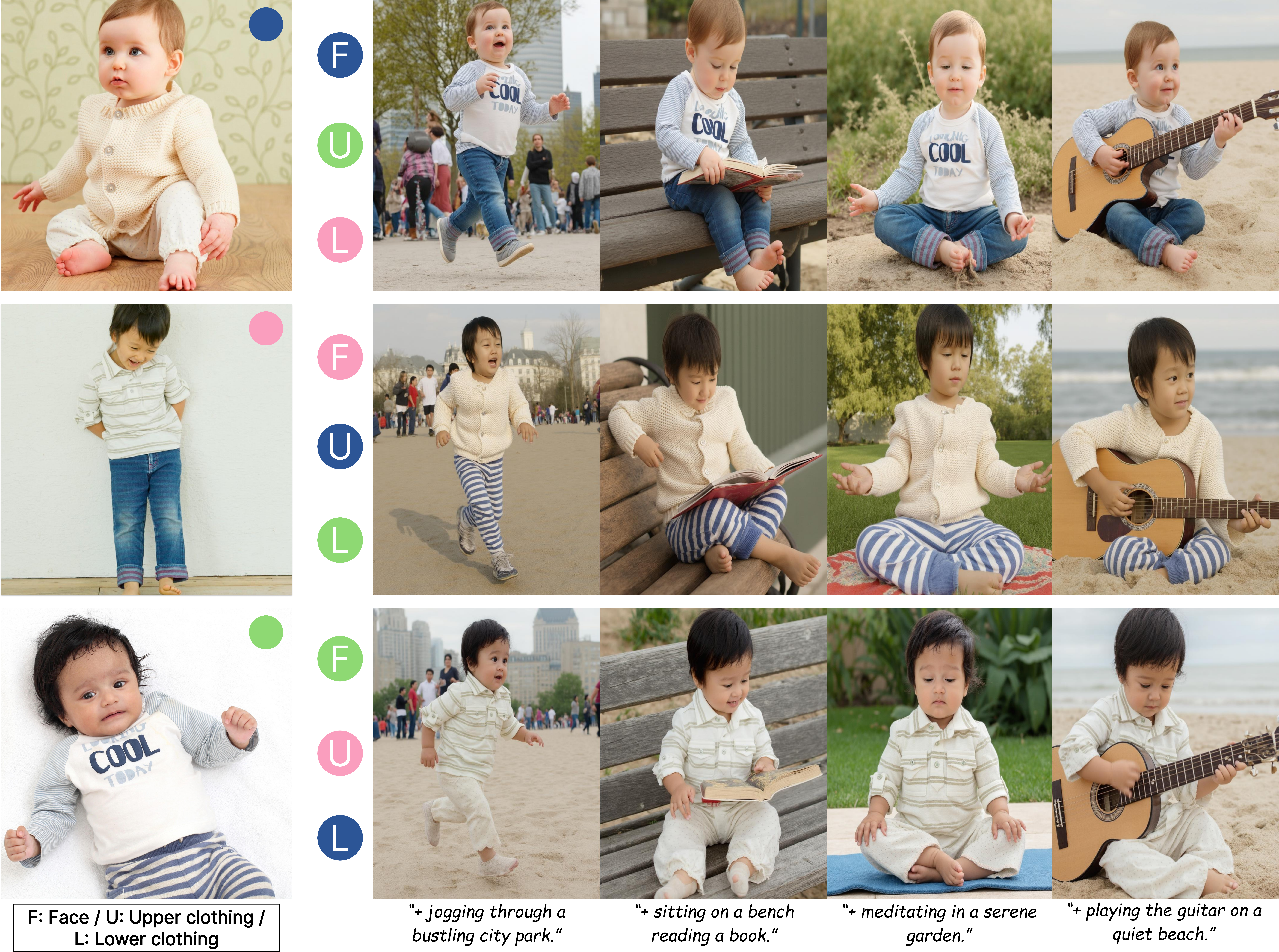}
    \vspace{-0.2cm}
    \caption{\textbf{Qualitative results.} To the left, three individual images are shown, and in each image, the face, upper, and lower clothing are composited to generate four text prompts on the right. Our methodology accurately reflects fine details such as facial features and text of the clothing.}
    \vspace{-0.3cm}
    \label{fig:qual_test}
\end{figure*}

\subsection{Selective Subject Region Loss}
While reconstructing the entire image allows the model to reference wardrobe subjects, repeated training can unintentionally force every subject to appear in the final output—even when partially occluded or omitted (\textit{e.g.}, someone sitting behind a desk). To address this issue, we propose a selective subject region loss. As shown in Fig.~\ref{fig:main}, we ensure that part-level subjects from the wardrobe region are not necessarily included in the canvas and can be referenced independently. To achieve this, we randomly drop each subject’s region from the target individual with probability $p_{\text{drop}}$. We only include the region corresponding to the selected subject in the loss function:

\begin{equation}
    \begin{split}
        M'_{i} &= \bigcup_{j=1}^{m} \mathds{1}(p_j < p_{\text{drop}}) M_{ij}, \quad p_j \sim \mathcal{U}(0,1) \\[6pt]
        \mathcal{L}_{\text{ssr}} &= \mathbb{E}_{t,\,p_t(\mathbf{z}^{\prime} \mid \boldsymbol{\epsilon}),\,p(\boldsymbol{\epsilon})}
        \bigl[\|M' \odot \mathbf{v}_\Theta(\mathbf{z}^{\prime}, t, \mathbf{m}) \\
        &\quad\quad\quad\quad\quad\quad\quad\quad - M' \odot \mathbf{u}_t(\mathbf{z}^{\prime} \mid \boldsymbol{\epsilon})\|^2 \bigr].
    \end{split}
\end{equation}

To ensure a clearer distinction in the effect of selectively dropping subjects, even when a reference subject is provided, we adopt a training strategy in which either the reconstruction loss or the proposed loss is stochastically selected at each training step. With probability $p_{\text{ssr}}$, we apply the selective subject region loss; otherwise, we use the original flow matching loss:
\begin{equation}
    \mathcal{L} =
    \begin{cases}
        \mathcal{L}_{\text{ssr}}, & \text{if } p < p_{\text{ssr}}, \quad p \sim \mathcal{U}(0,1), \\
        \mathcal{L}_{\text{rec}}, & \text{otherwise}.
    \end{cases}
\end{equation}


\begin{figure}[t!]
    \centering
    \includegraphics[width=0.9\linewidth]{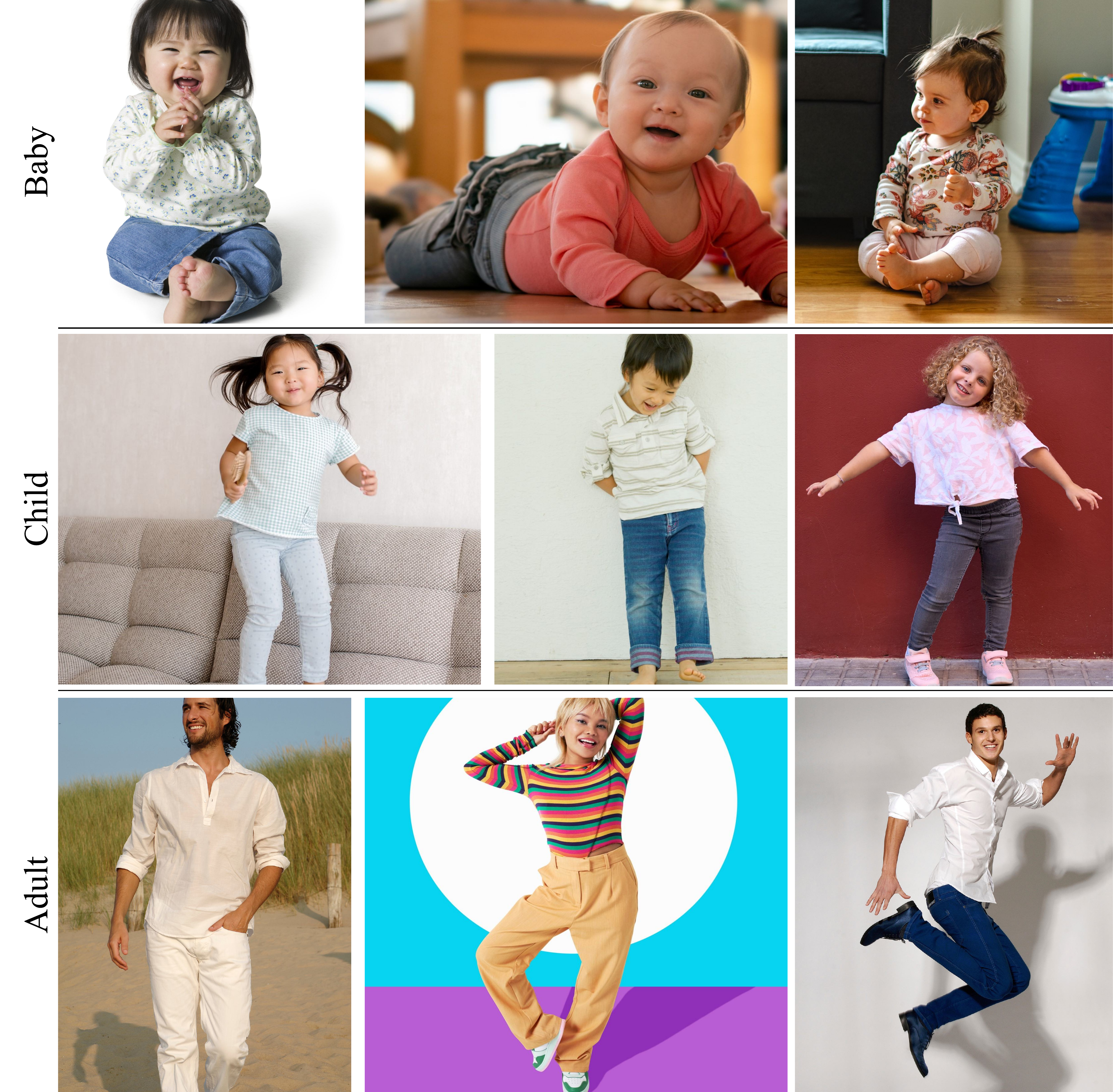}
    \caption{Examples of our Persona-36 dataset. We grouped participants by gender and divided them into three age categories—baby, child, and adult—collecting images from six individuals in each group.}
    \vspace{-0.3cm}
    \label{fig:dataset}
\end{figure}

\section{Experiment}
\noindent\textbf{Persona-36 Dataset.}
We introduce Persona-36, a dataset consisting of 36 unique identities collected from Getty Images\footnote{\url{https://www.gettyimages.com/}}. It includes six individuals per gender category (male and female) across three age groups: adult, child, and baby. Each identity is represented by 3 to 5 multi-view images captured in the same outfit. Fig.~\ref{fig:dataset} presents example images used for training and evaluation. 
To ensure a fair evaluation, we split the dataset into 24 and 12 identities for training and test respectively, maintaining no overlap between the two sets.

\noindent\textbf{Evaluation Metrics.}
For evaluation, we categorize identity combinations into the following three groups: subject composition from the same individual (set 1), subject composition from individuals of similar age (set 2), and subject composition across all age groups (set 3). From these three combination types, we created a total of nine identity combinations. Each combination was used to generate images from 30 prompts, with four images per prompt generated using different random seeds, resulting in a total of 1,080 generated images.

To quantitatively assess our approach and compare it with baseline methods, we employ two evaluation metrics: prompt similarity and identity similarity.
Prompt similarity measures the degree of correspondence between the input text prompt and the entire region of the generated image. We compute this using the standard CLIP similarity metric~\cite{radford2021learning}, specifically by calculating the cosine similarity between the normalized CLIP embeddings of the input prompt and the generated image. For prompt similarity, we focus primarily on comparing the background and action elements. To do this, we use a text prompt such as “a boy is taking photographs of a stunning sunset,” which includes only the background and action components for comparison. In our method, we compare the text prompt with the generated image in the canvas region.

Identity similarity evaluates how accurately compositional subjects are represented in the generated image, focusing on individual subject appearance rather than overall image similarity. To achieve this, we use a pre-trained human parsing model~\cite{li2020self} to extract subject-specific cropped images from the generated output. We then compute the average similarity of the extracted subject embeddings using DINO~\cite{caron2021emerging}, comparing them with their corresponding reference subjects.

\noindent\textbf{Implementation Details.}
We utilized the FLUX.1-Fill-dev\footnote{\url{https://huggingface.co/black-forest-labs/FLUX.1-Fill-dev}} model as the backbone of our method. The resolution of the entire generated image was set to $1024 \times 1024$, and in the experiments presented in this paper, we used three subjects, with the resolution of the canvas area set to $1024 \times 768$. For segmentation, we employed SCHP~\cite{li2020self} and SAM2~\cite{ravi2024sam}, and in cases where the segmentation maps for individual subjects were inaccurate, we manually labeled them. We trained our model using the AdamW optimizer with a learning rate of $5e^{-5}$, $\beta_1=0.9$, $\beta_2=0.999$, and a weight decay of $0.01$ for $5000$ iterations, optimizing both the flow matching loss and our proposed loss. We set the probability $p_{\text{ssr}}=0.5$ and the probability of dropping each subject to $p_{\text{drop}}=0.3$.

\noindent\textbf{Baselines.}
WP-LoRA provides two key advantages for part-level subject generation: (1) it enables training a single model across multiple individuals, and (2) it facilitates the generation of images for unseen individuals with minimal data, without requiring additional fine-tuning. To evaluate its performance, we compare WP-LoRA with baseline models that either learn single- or multi-subject representations or encode multiple subjects for generation without further fine-tuning for the new dataset.

For the single-subject model (DreamBooth), we set a specific individual as the subject (\textit{e.g.}, ``a sks man") for individual images. For multiple individuals (sets 2 and 3), we assign unique tokens to each subject and train the model jointly on all individuals. In contrast, multi-subject models (Break-A-Scene) are trained using multi-subject tokens for all cases. These per-subject optimization-based methods were trained on 9 different combinations of the training dataset. For PartCraft and Parts2Whole, we train on 24 individuals and evaluate the models on 9 training and 9 test combinations.

Models excluding DreamBooth require additional GPU memory due to the use of attention maps in the loss function or additional encoders, which limits their scalability to large vision models (LVMs) such as FLUX\footnote{\url{https://huggingface.co/black-forest-labs/FLUX.1-dev}}. To mitigate such limitations, we replaced the backbone weights with RealVision 3.0, a high-quality variation model capable of generating detailed human images.


\begin{figure*}[t!] 
    \centering
    \includegraphics[width=\linewidth]{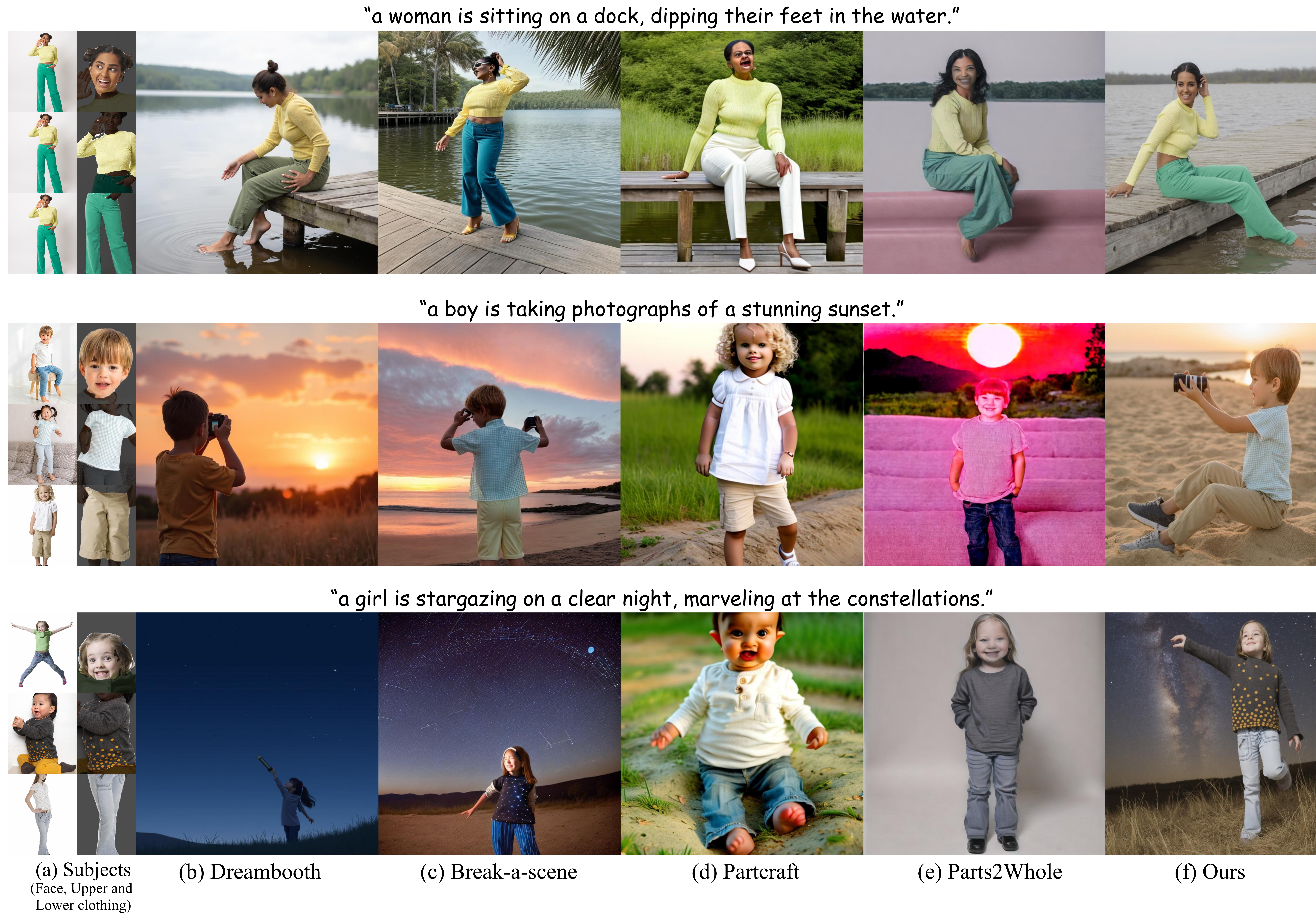}
    \vspace{-0.4cm}
    \caption{Qualitative results of training sets. Each row represents a newly generated individual composed of three subjects from the training dataset. The first row contains compositions from the same individual, while the second and third rows combine different individuals. For clarity, we visualize each subject alongside its original reference on the far left. Our method produces the most detailed results and accurately reflects the provided subjects. Notably, unlike DreamBooth and Break-A-Scene, our approach successfully generates results using a single trained model.}
    \vspace{-0.3cm}
    \label{fig:qual_train}
\end{figure*}

\begin{figure}[t!]
    \centering
    \includegraphics[width=\linewidth]{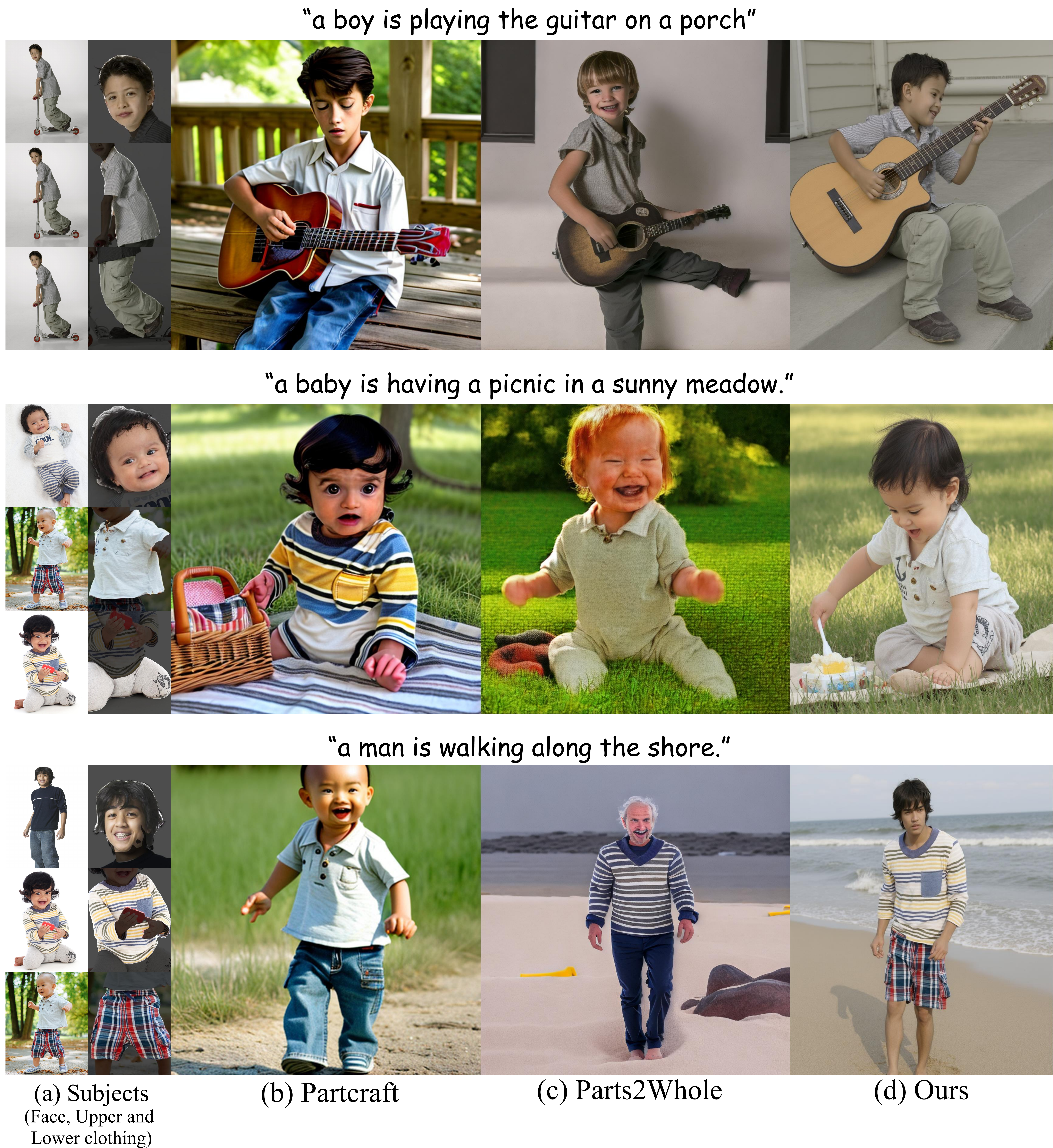}
    \vspace{-0.5cm}
    \caption{Qualitative results of test sets. Our model demonstrates high identity similarity even for individuals unseen during training.}
    \vspace{-0.3cm}
    \label{fig:qual_test}
\end{figure}

\begin{figure}[t!]
    \centering
    \includegraphics[width=\linewidth]{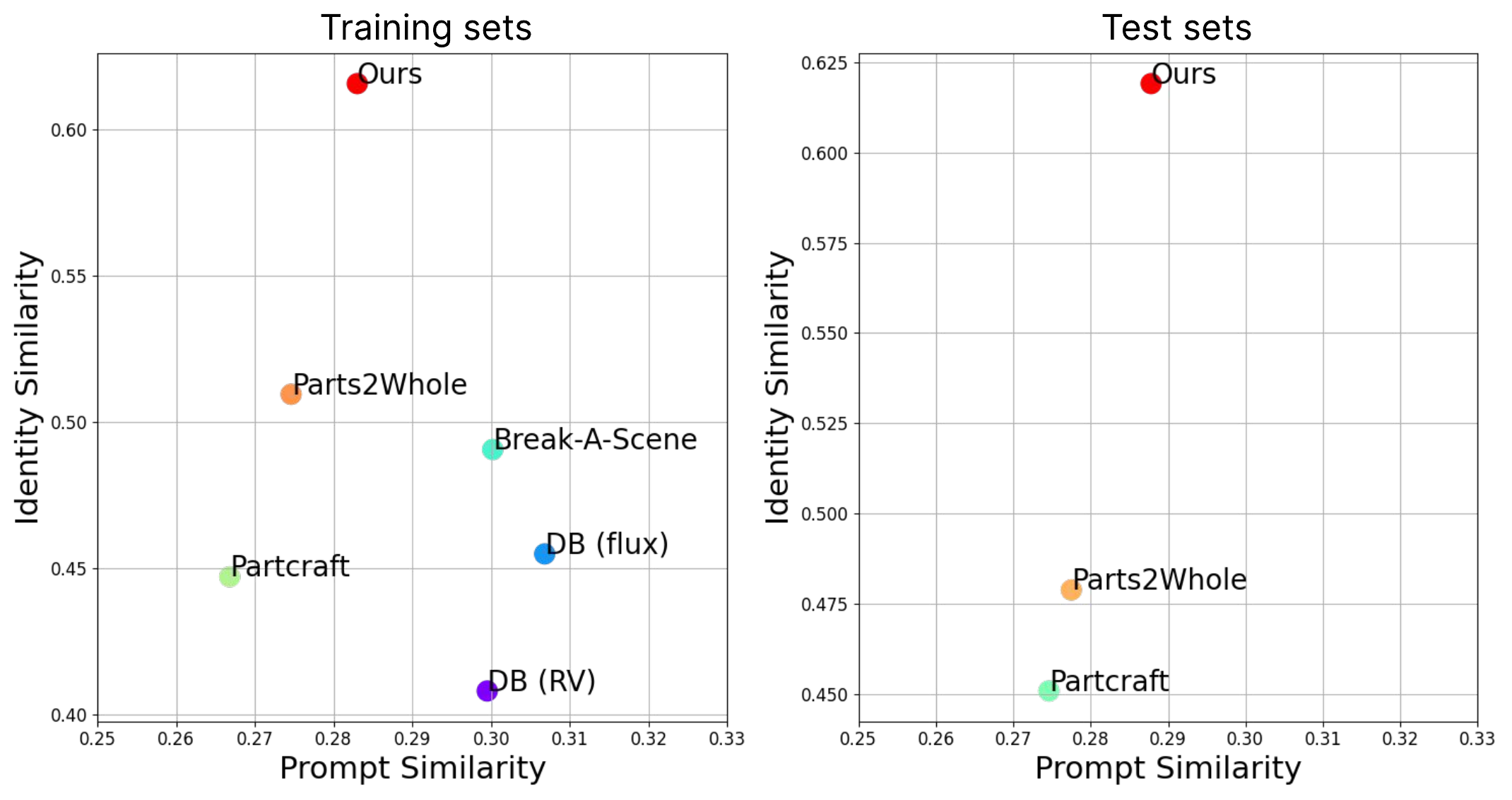}
    \vspace{-0.6cm}
    \caption{Quantitative results of training and test sets.}
    \vspace{-0.6cm}
    \label{fig:quan}
\end{figure}

\subsection{Comparative Evaluation}
Fig.~\ref{fig:qual_train} and Fig.~\ref{fig:qual_test} present qualitative results for 24 trained subjects and 12 test subjects, respectively. The rows in each figure correspond to set 1, set 2, and set 3.

As shown in the second and third rows of Fig.~5, DreamBooth, which is designed for single-subject learning, completely fails to preserve subject identity. Similarly, Break-A-Scene generates clothing and pants with incorrect patterns, as observed in the third row. PartCraft and Parts2Whole suffer from overfitting, failing to retain subject identity and showing poor text prompt adherence. These issues are are also shown in test datasets.

In contrast, our method, using a single trained model, accurately captures both text descriptions and subject identities across training and test sets. 
Fig.~\ref{fig:quan} quantitatively compares identity and prompt similarity across methods.

While our approach shows a slight decrease in prompt similarity compared to models that learn unique tokens (DreamBooth and Break-A-Scene), it achieves significantly higher identity similarity. 
Since preserving subject identity is the primary goal in generating novel outfit compositions, this highlights the effectiveness of our method. 
Moreover, our model outperforms PartCraft and Parts2Whole in both identity and prompt similarity, demonstrating superior generalization and robustness. 
These results confirm that referencing a fixed subject in the wardrobe region enables effective part-level composition, even with limited training data.

\begin{figure}[t!]
    \centering
    \includegraphics[width=\linewidth]{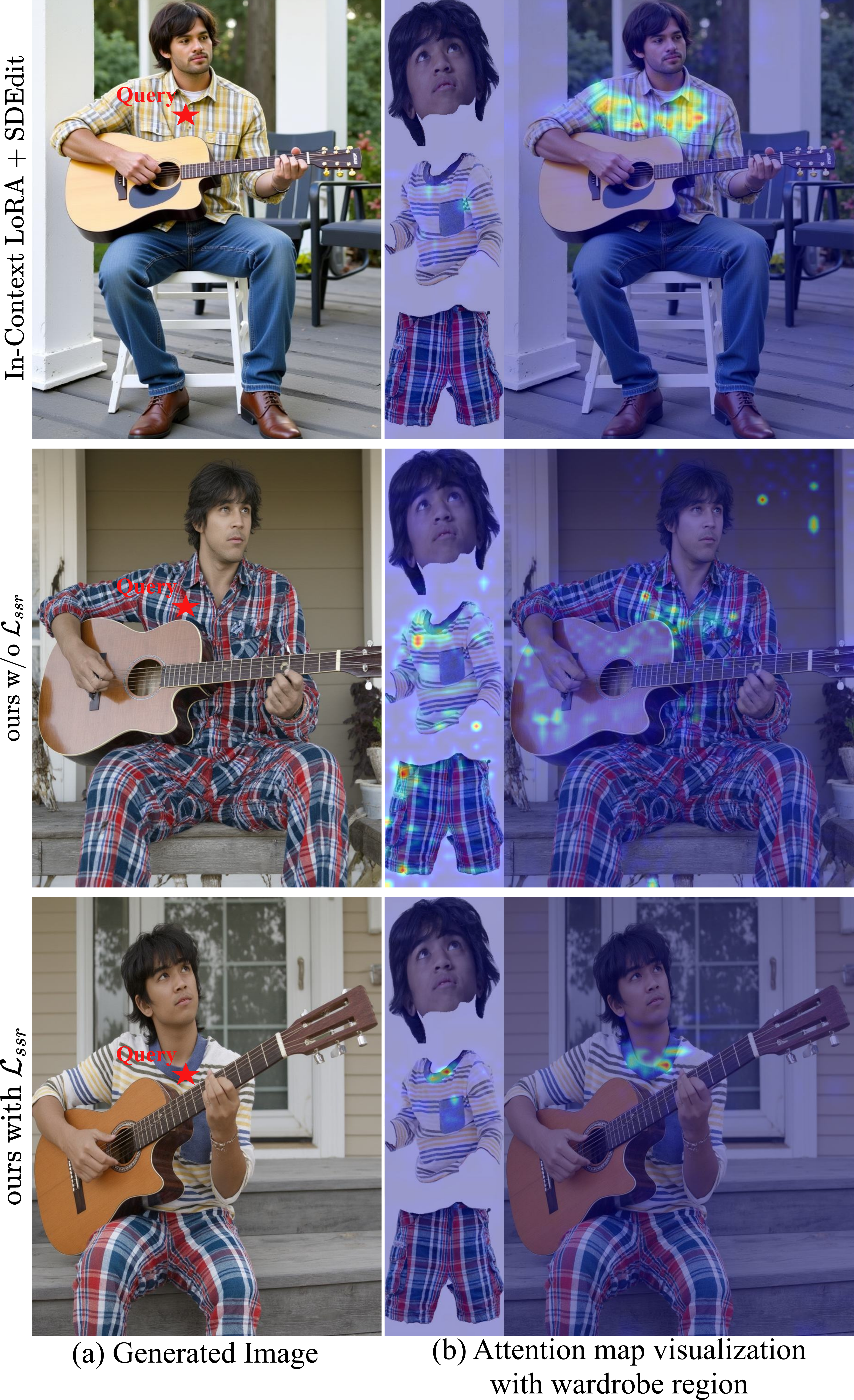}
    \vspace{-0.6cm}
    \caption{Visualization of the attention map of the query token for a specific subject in the generated image.}
    \vspace{-0.6cm}
    \label{fig:ablation}
\end{figure}

\begin{table}
\centering
\resizebox{\linewidth}{!}{
\setlength{\tabcolsep}{4pt} 
\begin{tabular}{lccc|ccc}
    \toprule
    & \multicolumn{3}{c}{\textbf{Training sets}} & \multicolumn{3}{c}{\textbf{Test sets}}  \\ 
    & \textbf{P.S.} & \textbf{I.S.} & \textbf{Avg.} & \textbf{P.S.} & \textbf{I.S.} & \textbf{Avg.} \\ 
    \midrule
    IC-LoRA + SDEdit & \textbf{0.2900} & 0.5043 & 0.3972 & \textbf{0.2912} & 0.5049 & 0.3980 \\
    Ours w/o $\mathcal{L}_{SSR}$  & 0.2846 & 0.5821 & 0.4334 & 0.2868 & 0.5878 & 0.4373 \\
    Ours with $\mathcal{L}_{SSR}$ & 0.2868 & \textbf{0.6077} & \textbf{0.4473} & 0.2885 & \textbf{0.6181} & \textbf{0.4533} \\
    \bottomrule
    \end{tabular}
}
\vspace{-0.3cm}
\caption{Quantitative comparison of ablation results. Here, P.S. and I.S. denote text prompt similarity and identity similarity, respectively.}
\vspace{-0.5cm}
\label{tab:ablation}
\end{table}


\noindent\textbf{Comparison with In-Context LoRA.}
Similar to our method, In-context LoRA (IC-LoRA)~\cite{huang2024context} generates consistent images by concatenating images with a consistent style or similar appearance, either vertically or horizontally, within the DIT model. 
To condition specific subjects, IC-LoRA adopts SDEdit~\cite{meng2021sdedit}, which replaces noised subject images at each inference step.
As shown in the first column of Fig.~\ref{fig:ablation} and Table~\ref{tab:ablation}, IC-LoRA still shows insufficient identity fidelity. 
We visualized self-attention maps for a query token (marked with an asterisk in the first column) located in the upper garments, taken from intermediate transformer layers. 
These maps were overlapped with the input wardrobe and canvas regions. 
The attention map in the top right reveals that IC-LoRA + SDEdit exhibits low activation between the queries and their corresponding subjects, highlighting its limitation in referencing subjects placed in specific regions.
Compared to IC-LoRA + SDEdit, which suffers from information loss in the conditioned subject (particularly during early inference steps) due to noise addition, our proposed method provides clean RGB images and demonstrates superior capability in referencing subjects.

\noindent\textbf{Effectiveness of $\mathcal{L}_{ssr}$.}
In Fig.~\ref{fig:ablation}, we demonstrate the effect of our proposed selective loss through attention map visualization. 
As shown in the second row, when challenging compositional scenarios arise such as occlusion caused by foreground objects (\textit{e.g.,} guitars), the model struggles to maintain appearance separation, resulting in attributes from lower garments inappropriately transferring to and blending with upper garments.
The attention map shows that activated keys for query tokens are sparsely distributed across reference upper and lower garments. 
In contrast, our proposed loss independently reconstructs each subject, resulting in precise attention to upper garments even in complex poses and occlusion scenarios, as shown in the bottom right. 
This leads to quantitative improvements in identity preservation, as demonstrated in Table~\ref{tab:ablation}.

\section{Conclusion}

We aim to address part-level controllable human image generation, where precise and consistent attributes are crucial. 
Different from the existing approaches that often demand large-scale training datasets, we introduce Wardrobe Polyptych LoRA, a model capable of controllable image generation trained with fewer than 100 images. 
WP-LoRA leverages the spatial layout of the generated image to reference conditioned subjects, functioning without additional parameters during inference and performing well even with limited data on unseen individuals.
Our approach opens up possibilities for efficient and scalable human image editing.

{
    \small
    \bibliographystyle{ieeenat_fullname}
    \bibliography{main}
}

\end{document}